\documentclass[twoside,twocolumn,10pt]{article}

\usepackage{wscg}           
\usepackage{soul}
\RequirePackage{ifpdf}
\ifpdf
 \RequirePackage[pdftex]{graphicx}
 \RequirePackage[pdftex]{color}
\else
 \RequirePackage[dvips,draft]{graphicx}
 \RequirePackage[dvips]{color}
\fi
\usepackage[polish,english]{babel}     
\usepackage{array}          
\usepackage[intlimits]{amsmath} 
\usepackage{hyphenat}      

\usepackage{stackengine}

\usepackage{comment}
\usepackage{subcaption}
\usepackage{hyperref}

\usepackage[style=alphabetic]{biblatex}
\DeclareLabelalphaTemplate{
  \labelelement{
    \field[final]{shorthand}
    \field{label}
    \field[strwidth=3,strside=left,names=1]{labelname}
  }
  \labelelement{
    \field[strwidth=2,strside=right]{year}    
  }
}

\graphicspath{{figures/}} 
\usepackage{nopageno}       
\usepackage[switch,columnwise]{lineno}

\title{FastRIFE: Optimization of Real-Time Intermediate Flow Estimation for Video Frame Interpolation}

\author{
\parbox{0.5\textwidth}{\centering
Malwina Kubas\\[1mm]
Warsaw University of Technology\\
Faculty of Electrical Engineering\\
ul. Koszykowa 75\\
00-662 Warszawa, Poland\\[1mm]
malwina.kubas.stud@pw.edu.pl
}
\hspace{0.05\textwidth}
\parbox{0.5\textwidth}{\centering
Grzegorz Sarwas\\[1mm]
Warsaw University of Technology\\
Faculty of Electrical Engineering\\
ul. Koszykowa 75\\
00-662 Warszawa, Poland\\[1mm]
grzegorz.sarwas@pw.edu.pl
}
}

\usepackage{url}
\makeatletter
\def\Uslash{\mathbin{\mathchar`\/}\@ifnextchar{/}{\kern-.15em}{}}
\g@addto@macro\UrlSpecials{\do \/ {\Uslash}}
\def\Ucolon{\mathbin{\mathchar`:}\@ifnextchar{/}{\kern-.1em}{}}
\g@addto@macro\UrlSpecials{\do : {\Ucolon}}
\makeatother

\begin{document}

\twocolumn[{\csname @twocolumnfalse\endcsname

\maketitle  

\begin{abstract}
\noindent
The problem of video inter-frame interpolation is an essential task in the field of image processing. Correctly increasing the number of frames in the recording while maintaining smooth movement allows to improve the quality of played video sequence, enables more effective compression and creating a slow-motion recording. This paper proposes the FastRIFE algorithm, which is some speed improvement of the RIFE (Real-Time Intermediate Flow Estimation) model. The novel method was examined and compared with other recently published algorithms. All source codes are available at: \\ \textsl{
 	\url{https://gitlab.com/malwinq/interpolation-of-images-for-slow-motion-videos} .}
\end{abstract}

\subsection*{Keywords}
Video Interpolation, Flow Estimation, Slow-Motion

\vspace*{1.0\baselineskip}
}]

\section{Introduction}


Video frame interpolation is one of the most important issues in the field of image processing. Correctly reproduced or multiplied inter-frame information allows its use in a whole range of problems, from video compression \cite{Wu18}, through improving the quality of records, to generating slow-motion videos \cite{Men20} or even view synthesis \cite{Flynn16}. Mid-frame interpolation performed in real-time on high-resolution images also increases the image's smoothness in video games or live broadcasts. With fast and accurate algorithms, we can reduce the costs associated with the construction of high-speed video cameras or provide services to users with limited hardware or transmission resources.

Algorithms of this class should deal with complex, non-linear motion models, as well as with changing parameters of the real vision scene, such as shadows, changing the color temperature, or brightness. Conventional approaches are based on motion prediction and compensation \cite{Haan93, Bao18}, which are used in various display devices \cite{Wu15}.

\begin{figure}[b!]
    \noindent\fbox{%
        \parbox{\dimexpr\linewidth-2\fboxsep-2\fboxrule\relax}{%
           \small{Permission to make digital or hard copies of all or part of this work for personal or classroom use is granted without fee provided that copies are not made or distributed for profit or commercial advantage and that copies bear this notice and the full citation on the first page. To copy otherwise, or republish, to post on servers or to redistribute to lists, requires prior specific permission and/or a fee.}
        }%
    }
\end{figure}

Motion estimation is used to determine motion vectors in a block or pixel pattern between two frames. Block-based methods \cite{Haan93} assume that pixels in a block have the same inter-frame shift and use search strategies \cite{Zhu00, Gao00}, and selection criteria \cite{Haan93, Wang10} to obtain the optimal motion vector. Other methods of movement estimation are optical flow algorithms that estimate the motion vector for each pixel separately, which makes them more computationally expensive. In recent years there has been noticed significant progress in this area. New algorithms have been developed based on the optimization of variance \cite{Brox04}, searching for the nearest neighbors \cite{Chen13}, filtering the size of costs \cite{Xu17} and deep convolutional neural networks (CNN) \cite{Ranjan17,rel2}. However, very often, the quality of the obtained interpolations is burdened with an increase in the required computing resources and often limits their use in the case of users working on standard equipment or mobile phones.

\begin{figure*}[!ht]
	\centering
	\includegraphics[width=0.99\textwidth]{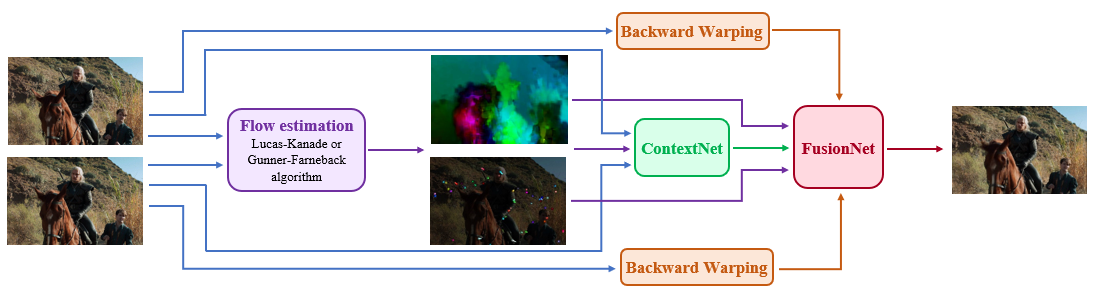}
	\caption{Architecture of the proposed video frame interpolation algorithm. We estimate the optical flow with analytical methods and use fine-tuned neural network models from RIFE: ContextNet and FusionNet.}
	\label{structure}
\end{figure*}

Flow-based  video frame interpolation algorithms achieve very good results \cite{superslomo, Niklaus18, dain, Xue19, rife}. These solutions include two steps:  warping the input frames according to approximated optical flows and fusing and refining the  warped  frames  using  CNN. Flow-based algorithms can be divided into forward warping methods and more often used back-warping methods. A common operation of these methods is to calculate bidirectional optical flow to generate an intermediate flow. These operations require a large number of computing resources and are not suitable for real-time scenarios.

This work aims to create an efficient video frame interpolation model, which can be implemented on many devices and run in real-time. 
Authors will compare state-of-the-art techniques, best in image interpolation field at the moment, and propose an improvement in terms of execution time in one of them. Algorithms which will be analyzed and compared are: MEMC-Net \cite{rel4}, SepConv \cite{sepconv}, Softmax Splatting \cite{softmax}, DAIN \cite{dain} and RIFE \cite{rife}, all using different approaches. 

\section{Related Work}

The topic of video frame interpolation has been widely discussed in the research papers \cite{rel1,rel2,sepconv,rel4,superslomo}. The first groundbreaking method was proposed in \cite{first}. Long \textit{et al.} were the first to use the convolutional neural network, which aimed to estimate the result frame directly. Next generations of image interpolation algorithms brought significantly better results, each using more extensive neural network structures.

Some algorithms, like SepConv \cite{sepconv}, were built and trained end-to-end using only one pass through the encoder-decoder structure. However the majority of methods use an additional sub-networks, of which the most important is the optical flow estimation. Having the information about pixels transition allows generating more accurate results and boost benchmark metrics. This approach was used e.g. in the MEMC-Net method, which combines both motion estimation, motion compensation and post-processing by four sub-networks \cite{rel4}. 

Bao \textit{et al.} composed an improvement of MEMC-Net model called DAIN (Depth-Aware INterpolation) \cite{dain}, where the information  of pixels depth was used, improving dealing with the problems related to occlusion and large object motion by sampling closer objects better. Depth estimation is one of the most challenging tasks in the image processing field of study, so the whole process takes a noticeable amount of time. Authors used a fine-tuned model called hourglass \cite{dain_depth} network learned on the MegaDepth dataset \cite{megadepth}. DAIN method estimates the bi-directional optical flow using an off-the-shelf solution called PWC-Net \cite{Xue19}.

Another excellent flow-based method is Softmax Splatting \cite{softmax}. Niklaus and Liu proposed a solution of forward warping operator, which works conversely to backward warping and is used in most methods. Softmax Splatting uses an importance mask and weights all pixels in the input frame. Authors used fine-tuned PWC-Net for estimating optical flow, similarly to DAIN.

One of the newest algorithm is RIFE - Real-Time Intermediate Flow Estimation \cite{rife}, which is able to achieve comparable results on standard benchmarks when compared to previous methods but also works significantly faster. More details about this solution are presented in the next section.

\begin{figure*}[!ht]
\captionsetup[subfigure]{justification=centering}
\begin{subfigure}{.16\textwidth}
  \centering
  \includegraphics[width=\textwidth]{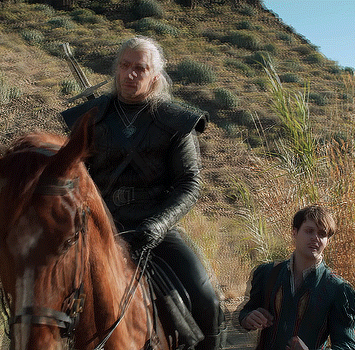}
  \caption*{Overlayed \\ inputs}
  \label{fig3:1}
\end{subfigure}
\begin{subfigure}{.16\textwidth}
  \centering
  \includegraphics[width=\textwidth]{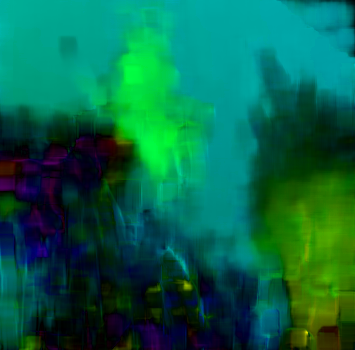}
  \caption*{Optical flow \\ by GF method}
  \label{fig3:2}
\end{subfigure}
\begin{subfigure}{.16\textwidth}
  \centering
  \includegraphics[width=\textwidth]{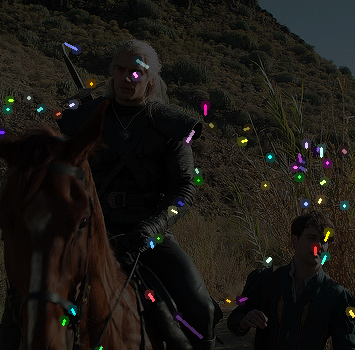}
  \caption*{Optical flow \\ by LK method}
  \label{fig3:3}
\end{subfigure}
\begin{subfigure}{.16\textwidth}
  \centering
  \includegraphics[width=\textwidth]{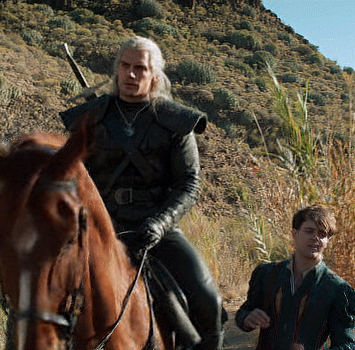}
  \caption*{Interpolated frame \\ with GF method}
  \label{fig3:4}
\end{subfigure}
\begin{subfigure}{.16\textwidth}
  \centering
  \includegraphics[width=\textwidth]{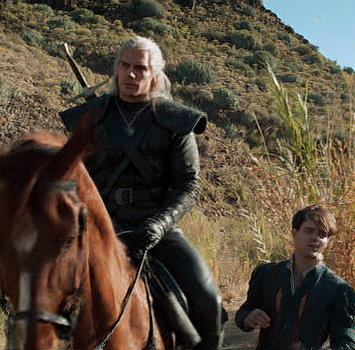}
  \caption*{Interpolated frame \\ with LK method}
  \label{fig3:5}
\end{subfigure}
\begin{subfigure}{.16\textwidth}
  \centering
  \includegraphics[width=\textwidth]{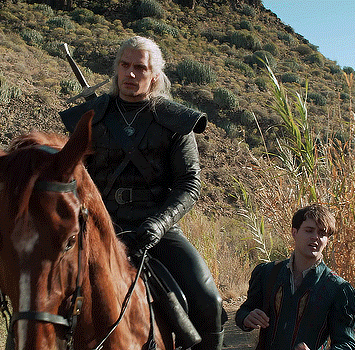}
  \caption*{Ground-truth \\ frame}
  \label{fig3:6}
\end{subfigure}
\vspace{4px}

\caption{Example of video frame interpolation results. Authors propose a simplified RIFE model with optical flow estimated using analytical methods: Gunnar-Farneb\"ack and Lucas-Kanade.}
\label{fig:example}
\end{figure*}

\section{Proposed solution}
Most state-of-the-art algorithms generate intermediate frames by combining bi-directional optical flows and additional networks, e.g. estimating depth, which makes these methods unable to real-time work. Huang \textit{et al.} proposed in RIFE \cite{rife} a solution of simple neural network for estimating optical flows called IFNet and a network for generating interpolated frames called FusionNet.

The IFNet comprises three IFBlocks, each built with six ResNet modules and operating on increasing image resolutions. The output flow is very good quality and runs six times faster than the PWCNet and 20 times faster than the LiteFlowNet \cite{Hui18}. After estimating the flow, coarse reconstructions are generated by backward warping and then the fusion process is performed. The process includes context extraction and the FusionNet with an architecture similar to U-Net, both consisting four ResNet blocks. 

In this paper authors will analyze the RIFE model with different module for estimating optical flow. Although the IFNet gives excellent results and runs very fast, authors wanted to test analytical methods and compare the results of such a simplified model in terms of runtime and quality. This change can speed up the algorithm even more, making the whole process capable of running real-time on many more devices. 

\begin{figure}[!ht]
	\centering
	\includegraphics[scale=0.48]{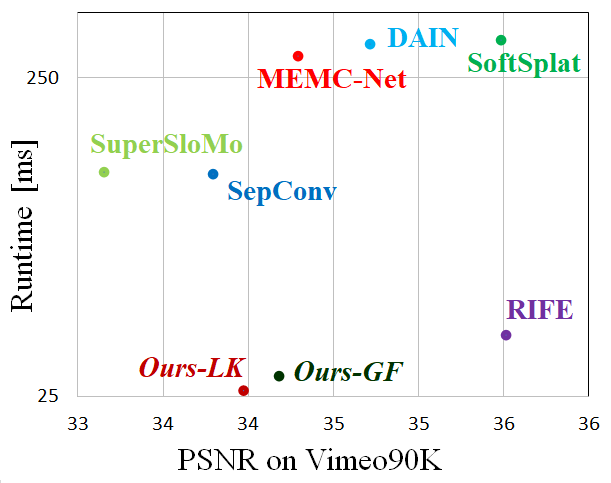}
	\caption{Speed and accuracy trade-off measured on RTX 2080 Ti GPU. Our models compared with previous frame interpolation methods: RIFE \cite{rife}, Softmax Splatting \cite{softmax}, DAIN \cite{dain}, MEMC-Net \cite{rel4}, SuperSloMo \cite{superslomo} and SepConv \cite{sepconv}. Comparison done on Vimeo90K test dataset. Please note the logarithmic runtime scale.}
	\label{tradeoff}
\end{figure}

Authors replaced the IFNet part with analytical methods described below. Then, new FusionNet and ContextNet models were fine-tuned on the proposed solution. The full architecture is shown in Figure \ref{structure}. Authors decided to test two of the most widely used algorithms, one generating dense flow and one which estimates sparse results. The example of optical flows and frame interpolation results are shown in Figure \ref{fig:example} and the speed/accuracy trade-off is presented in Figure \ref{tradeoff}.

\subsection{Gunnar-Farneb\"ack algorithm}
The Gunnar-Farneb\"ack method \cite{dense} gives dense results, which means that flow values are generated for every pixel. The algorithm detects changes in pixel intensity between two images using polynomial expansions and highlights pixels with the most significant changes.

The idea depends on an approximation of neighbourhood pixels in both frames by quadratic polynomials. Pixels' displacements are estimated from the differentiation of transforms of polynomials under translations. The algorithm is implemented by hierarchical convolutions what makes the process very efficient. 

After the tests authors set the parameters of the Gunnar-Farneb\"ack (GF) method as follows:
\begin{itemize}
	\item scale of image pyramids: 0.2,
	\item number of pyramid layers: 3,
	\item size of the pixel neighborhood: 5,
	\item size of averaging windows: $15\times15$.
\end{itemize}
The above sets give the best runtime to accuracy trade-off.

\subsection{Lucas-Kanade algorithm}
Dense methods can give very accurate results. However, sometimes the speed to accurate ratio might be more favorable for sparse methods. These algorithms calculate optical flow only for some number for pixels, extracted by the detector as feature points. To feed the results to the FusionNet it was needed to change the values from sparse to dense form.

Authors used Lucas-Kanade (LK) algorithm, which is the most commonly used analytical method for generating optical flow. It takes a $3\times3$ patch around the point and assumes all nine pixels will have the same flow. The solution is calculated with the least squares method. It is proven that the Lucas-Kanade works better with corner points, so for feature points detections authors used the Shi-Tomasi algorithm. The function calculates corner quality measure in each pixel's region using minimum eigenvalues and Harris Corner Detection.

Parameters which were used in the Lucas-Kanade algorithm are:
\begin{itemize}
	\item size of search window: $15\times15$,
	\item number of pyramids: 1 (single level),
	\item termination criteria: epsilon and criteria count,
\end{itemize}
and for the Shi-Tomasi detector:
\begin{itemize}
	\item maximum number of corners: 100,
	\item minimum accepted quality of corners: 0.1,
	\item minimum possible distance between the corners: 10,
	\item size of computing block: 7.
\end{itemize}

\subsection{Runtime}
Runtime comparison can be found in Table \ref{tab:runtime}. IFNet works much faster than any other state-of-the-art method, but anyway optical flow can be obtained by analytical methods in less time. One of the important issues is that most of the standard flow methods have to run the estimation process twice if a bi-directional flow is desired. The results were measured again on RTX 2080 Ti GPU.

\begin{table}[ht]
	\centering
	\begin{tabular}{|l|l|}
	    \hline
		Method & Runtime [ms] \\
		\hline
		PWCNet & 125 \\
		LiteFlowNet & 370 \\
		IFNet & 34 \\
		\textbf{Ours - GF} & \textbf{9} \\
		\textbf{Ours - LK} & \textbf{7} \\
        \hline
	\end{tabular}
	\caption{Runtime comparison of optical flow estimation algorithms}
	\label{tab:runtime}
\end{table}

\section{Experiments}
This section will provide evaluation results, comparing the last and best methods with our proposal. Comparison is made based on quantitative and visual results using common metrics and datasets as well as algorithms runtime.

\subsection{Learning strategy}
Our solution was trained on Vimeo90K dataset with optical flow labels generated by the LiteFlowNet \cite{Hui18} in PyTorch version \cite{torch-liteflownet}. Authors used the AdamW optimizer, the same as used in RIFE \cite{rife}.

\subsubsection{Loss function}
To address the problem of training the IFNet module, Huang \textit{et al.} proposed a solution of leakage distillation schema, which compares the network result with the output of the pre-trained optical flow estimation method. The training loss is a linear combination of reconstruction loss $L_{rec}$, census loss $L_{cen}$ and leakage distillation loss $L_{dis}$. Our solution was trained using the same loss function:
\begin{equation}
 L = L_{rec} + L_{cen} + \lambda L_{dis},    
\end{equation}
with the weight of leakage distillation loss 10 times smaller than the two others ($\lambda = 0.1$).

\subsubsection{Training dataset}
Vimeo90K is a video dataset containing 73,171 frame triplets with $448 \times 256$ pixels image resolution \cite{Xue19}. This dataset was used as a training set in all compared methods.

\subsubsection{Computational efficiency}
All analyzed algorithms are implemented in PyTorch \cite{pytorch} using CUDA, cuDNN, and CuPy \cite{cuda}. The proposed solutions was implemented using OpenCV functions, and RIFE model was fine-tuned on Nvidia RTX 2080 Ti GPU with 11GB of RAM for 150 epochs which took 15 days to converge.

\subsection{Evaluation datasets}
For evaluation, the following publicly available datasets have been used:

\subsubsection{Middlebury}
Middlebury contains two subsets: Other with ground truth interpolation results and Evaluation, which output frames are not publicly available \cite{middlebury}. It is the most widely used dataset for testing image interpolation algorithms. The resolution of frames is $640 \times 480$ pixels.

\subsubsection{UCF101}
UCF101 provides 379 frame triplets from action videos with 101 categories \cite{ucf} \cite{ucf2}. Images resolution is $256 \times 256$ pixels.

\subsubsection{Vimeo90K}
Vimeo90K was used for training, but it contains also the test subset - 3,782 triplets with $448 \times 256$ pixels image resolution \cite{Xue19}.

\begin{figure*}[!ht]
	\centering
	\includegraphics[scale=0.75]{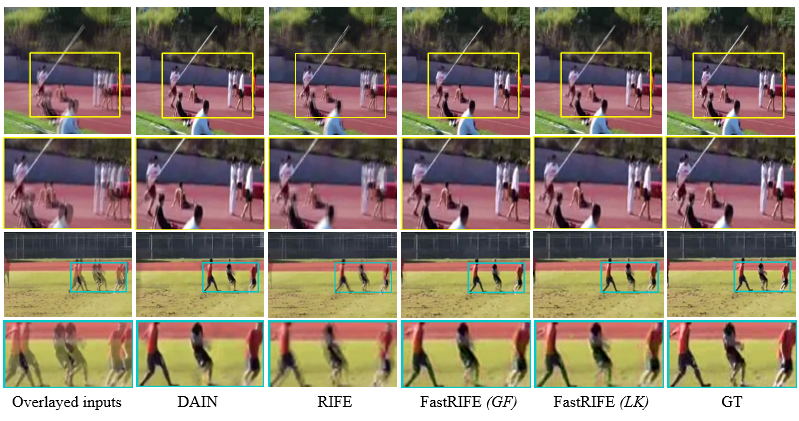}
	\caption{Qualitative comparison on UCF101 dataset}
	\label{visual_compare}
\end{figure*}

\subsection{Metrics}
The following metrics have evaluated each algorithm.
PSNR (Peak Signal-to-Noise Ratio):
\begin{equation}
    PSNR(i,j) = 10 \cdot \log_{10} \frac{255^2}{MSE(i,j)}. 
\end{equation}
SSIM (Structural Similarity, $l$ - luminance, $c$ - contrast and $s$ - structure):
\begin{equation}
 SSIM(i,j) = l(i,j) \cdot c(i,j) \cdot s(i,j).    
\end{equation}
IE (Interpolation Error) is the arithmetic average of a difference between the interpolated image and the ground truth frame:
\begin{equation}
IE(i,j) = \overline{|i - j|}, 
\end{equation}
where $i$ is interpolated frame and $j$ is the ground truth image \cite{metrics}. On UCF101 and Vimeo90K datasets we used PSNR and SSIM (higher values mean better results), on Middlebury Evaluation and Other we evaluated IE (lower values mean better results). This set of benchmarks is used in the majority of frame interpolation articles and datasets websites.

\begin{table*}[ht]
	\centering
	\begin{tabular}{|l|c|ll|ll|c|}
	    \hline
		{Method} & {Middlebury} & \multicolumn{2}{c|}{UCF101} &  \multicolumn{2}{c|}{Vimeo90K} & {Parameters} \\
		{} & IE & PSNR & SSIM & PSNR & SSIM & (million) \\
		\hline
        SepConv & 2.27 & 34.78 & 0.967 & 33.79 & 0.970 & 21.6 \\
        MEMC-Net & 2.12 & 35.01 & \color{blue}{\ul{0.968}} & 34.29 & 0.970 & 70.3 \\
        DAIN & 2.04 & 34.99 & \color{blue}{\ul{0.968}} & 34.71 & \color{blue}{\ul{0.976}} & 24.0 \\
		SoftSplat & \color{red}{1.81} & \color{blue}{\ul{35.10}} & 0.948 & \color{blue}{\ul{35.48}} & 0.964 & 7.7 \\
		RIFE & \color{blue}{\ul{1.96}} & \color{red}{35.25} & \color{red}{0.969} & \color{red}{35.51} & \color{red}{0.978} & 9.8 \\
		\textbf{Ours: FastRIFE - GF} & \textbf{2.89} & \textbf{34.84} & \textbf{\color{blue}{\ul{0.968}}} & \textbf{34.18} & \textbf{0.968} & \textbf{4.1} \\
		\textbf{Ours: FastRIFE - LK} & \textbf{2.91} & \textbf{34.26} & \textbf{0.967} & \textbf{33.97} & \textbf{0.967} & \textbf{4.1} \\
	\hline
	\end{tabular}
	\caption{Quantitative comparison of described methods on Middlebury Other, UCF101 and Vimeo90K test}
	\label{tab:test1}
\end{table*}

\subsection{Quantitative comparison}
Table \ref{tab:test1} shows a comparison of the results obtained on datasets: Middlebury Other, UCF101 and Vimeo90K, \color{red}{red} \color{black}{marks the best performance,} \color{blue}{\ul{blue}} \color{black}{marks second best score.} Results of evaluation on Middlebury benchmark are not available yet.

The proposed model performs comparably to other methods, scoring very similar results in both PSNR and SSIM. The most significant gap is between the results of Interpolation Error in Middlebury Other dataset. FastRIFE generates better values than the SepConv model in UCF101 and Vimeo90K benchmarks and worse results than other algorithms, but a significant reduction is shown on the parameters column. 

Evaluation of both optical flow methods gives similar results. However, the model which uses the Gunnar-Farneb\"ack algorithm performs slightly better in every benchmark, which means that the FusionNet behaves better with dense flow estimation.

\begin{figure}[ht]
	\centering
	\includegraphics[width=0.45\textwidth]{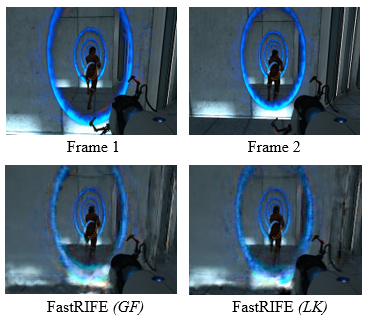}
	\caption{Visual comparison on HD frames}
	\label{game}
\end{figure}

\begin{table*}[ht]
	\centering
	\begin{tabular}{|l|c|c|}
	    \hline
		Method & Inference time on & Time of learning \\
		 & $448 \times 256$ image [ms] & one epoch [h] \\
		\hline
		SepConv & 125 & - \\
		MEMC-Net & 293 & - \\
		SoftSplat & 329 & - \\
		DAIN & 320 & 9 \\
		RIFE & 39 & 4.5 \\
		\textbf{Ours: FastRIFE - GF} & \textbf{29} & \textbf{2.4} \\
		\textbf{Ours: FastRIFE - LK} & \textbf{26} & \textbf{2.4} \\
        \hline
	\end{tabular}
	\caption{Runtime comparison of described methods tested on $448 \times 256$ image (device: RTX 2080 Ti GPU)}
	\label{tab:runtime2}
\end{table*}

\subsection{Visual Comparison}

Figure \ref{visual_compare} shows examples of frame estimation with comparison made on the DAIN and RIFE methods. These images show dynamic scenes, which are more challenging to analyze for frames interpolation methods. The resolution of images is small ($256\times256$), but both GF and LK methods generate acceptable results, similar to DAIN. The worst smudging is shown on RIFE results, which works better with high-resolution videos.

The results of high-resolution image interpolation are shown in Figure \ref{game}. Both FastRIFE models generate very blurry frames. It is probably caused by including only small resolution videos in the training process and setting constant parameters when calculating optical flow. The issue could be repaired as future work. Other methods: DAIN and RIFE work well with HD resolution frames.

The advantage of the proposed models is the size of occupied memory. For most methods, analyzing HD images results in an out-of-memory error on our GPU, while RIFE allocates only 3\ GB and FastRIFE only 1.5 GB of RAM. 

\begin{figure}[ht]
\centering
\begin{subfigure}{.235\textwidth}
  \centering
  \includegraphics[width=1\textwidth]{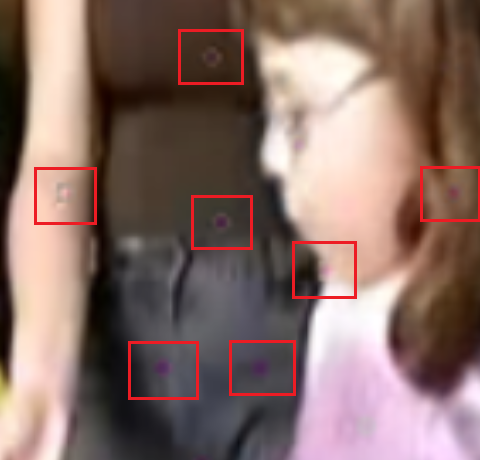}
  \caption*{FastRIFE - LK}
  \label{fig:issue1}
\end{subfigure}
\begin{subfigure}{.235\textwidth}
  \centering
  \includegraphics[width=0.95\textwidth]{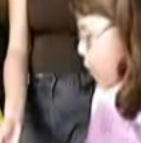}
  \caption*{FastRIFE - GF}
  \label{fig:issue2}
\end{subfigure}
\vspace{10px}
\caption{Sparse optical flow issue. Sometimes output frames suffers from spots visible in zoom}
\label{fig:issue}
\end{figure}

One important issue related to using the sparse optical flow method (Lucas-Kanade) is overlapping feature points as visible spots, as shown in Figure \ref{fig:issue}. The spots are noticeable after zooming small resolution images. Such a problem can be reported only for a model with the LK method. FastRIFE which uses the Gunnar-Farneb\"ack algorithm is this issue free.

\subsection{Runtime Comparison}

The usage of analytical methods and simplified RIFE model caused that FastRIFE performs favourably faster than any other state-of-the-art algorithm. Our models have 4 million parameters, which is the smallest neural network structure from all video frame interpolation methods. The runtime comparison is shown in Table \ref{tab:runtime2}.

Compared to any other model, except for RIFE, our solution runs up to 10 times faster. FastRIFE has slightly better benchmark results from SepConv, but runtime has been compressed from 125 to less than 30 milliseconds. The replacement of the optical flow module in RIFE saved 25\% of the time needed to interpolate frames. As shown in Table \ref{tab:runtime2}, also the time of learning one epoch was significantly reduced.

\section{Conclusion}

In this paper, we proposed a solution to RIFE model simplification in terms of optical flow estimation. FastRIFE uses well-known analytical methods instead of additional neural network modules, which results in excellent runtime improvement with an acceptable slight drop in the quality of output frames. 

Another advantage of model compression is the reduction of memory usage. Thanks to the small number of parameters, FastRIFE is able to generate output frames with the allocation of max 1.5 GB of RAM in the case of HD videos. Low memory consumption with short execution time makes the model possible to implement on many devices, including mobile phones.

The FastRIFE algorithm was analyzed with two method types for estimating optical flow: sparse (Lucas-Kanade) and dense (Gunnar-Farneb\"ack). Sparse flow performs faster, but better-quality results were always generated with the GF method. FastRIFE works appropriately in small resolution images but introduces prominently blurry regions when interpolating HD videos.

Future work that may be performed includes the improvement of the FastRIFE model in order to obtain better results on high resolution videos. It can probably be achieved by including HD images in the training dataset or by adjusting optical flow algorithm parameters on the fly based on video resolution or other criteria. Our proposed method can result in future research on increasing the efficiency of video frame interpolation algorithms. 

\printbibliography

\end{document}